\documentclass[sigconf, screen]{acmart}

\usepackage{color}
\usepackage{amsmath}
\usepackage{bm}

\usepackage{graphicx}
\usepackage{colortbl}
\usepackage{xcolor}
\usepackage{multirow}
\usepackage{makecell} 

\usepackage{caption}
\usepackage{algorithm}
\usepackage{algpseudocode}
\usepackage{tikz}
\usetikzlibrary{arrows}
\usepackage{tabulary}
\usepackage{booktabs}

\AtBeginDocument{%
  }

\setcopyright{acmlicensed}
\copyrightyear{2025}
\acmYear{2025}
\setcopyright{acmlicensed}\acmConference[MM '25]{Proceedings of the 33rd ACM International Conference on Multimedia}{October 27--31, 2025}{Dublin, Ireland}
\acmBooktitle{Proceedings of the 33rd ACM International Conference on Multimedia (MM '25), October 27--31, 2025, Dublin, Ireland}
\acmDOI{xxxxxxxx}
\acmISBN{xxxxxxxxx}

\begin{document}

\title{3D Gaussian Splatting Data Compression with Mixture of Priors}

\author{Lei Liu}
\affiliation{
  \institution{The University of Hong Kong}
  \city{Hong Kong SAR}
  \country{China}}
\email{liulei95@hku.hk}

\author{Zhenghao Chen$^*$}
\affiliation{%
  \institution{The University of Newcastle}
  \city{Newcastle}
  \country{Australia}}
\email{zhenghao.chen@newcastle.edu.au}

\author{Dong Xu$^*$}
\affiliation{
  \institution{The University of Hong Kong}
  \city{Hong Kong SAR}
  \country{China}}
\email{dongxu@hku.hk}

\renewcommand{\shortauthors}{Lei Liu, Zhenghao Chen, \& Dong Xu}

\begin{abstract}
  %
  %
3D Gaussian Splatting (3DGS) data compression is crucial for enabling efficient storage and transmission in 3D scene modeling. However, its development remains limited due to inadequate entropy models and suboptimal quantization strategies for both lossless and lossy compression scenarios, where existing methods have yet to 1) fully leverage hyperprior information to construct robust conditional entropy models, and 2) apply fine-grained, element-wise quantization strategies for improved compression granularity.
In this work, we propose a novel \textbf{Mixture of Priors (MoP)} strategy to simultaneously address these two challenges. Specifically, inspired by the Mixture-of-Experts (MoE) paradigm, our MoP approach processes hyperprior information through multiple lightweight MLPs to generate diverse prior features, which are subsequently integrated into the MoP feature via a gating mechanism. 
To enhance lossless compression, the resulting MoP feature is utilized as a hyperprior to improve conditional entropy modeling. 
Meanwhile, for lossy compression, we employ the MoP feature as guidance information in an element-wise quantization procedure, leveraging a prior-guided Coarse-to-Fine Quantization (C2FQ) strategy with a predefined quantization step value.
Specifically, we expand the quantization step value into a matrix and adaptively refine it from coarse to fine granularity, guided by the MoP feature, thereby obtaining a quantization step matrix that facilitates element-wise quantization.
%
Extensive experiments demonstrate that our proposed 3DGS data compression framework achieves state-of-the-art performance across multiple benchmarks, including Mip-NeRF360, BungeeNeRF, DeepBlending, and Tank\&Temples.
\end{abstract}





\maketitle

\let\thefootnote\relax\footnotetext{$^*$Zhenghao Chen and Dong Xu are the corresponding authors.}

\section{Introduction}
3D Gaussian Splatting (3DGS)~\cite{kerbl20233d} employs an explicit 3D representation using learnable Gaussian Splatting. Due to its high training efficiency and real-time rendering capabilities, it has rapidly emerged as a promising solution for high-quality novel view synthesis.
%
Despite its efficiency, 3DGS relies on a large number of Gaussians and their associated attributes to preserve visual fidelity, leading to significant storage and deployment overhead. This has motivated the development of dedicated compression techniques tailored to the unique characteristics of 3DGS.

Early approaches to compressing 3DGS primarily focus on reducing the parameter count and quantization to achieve lossy compression. These include clustering Gaussians into predefined codebooks via vector quantization~\cite{fan2023lightgaussian,lee2024compact,navaneet2023compact3d,niedermayr2024compressed}, or grouping them using anchor-based strategies~\cite{lu2024scaffold}.
However, these methods fall short in supporting lossless compression due to the absence of effective entropy coding techniques, and therefore cannot fully exploit the redundancy present in 3DGS representations.
To overcome this limitation, recent studies~\cite{wang2024contextgs,chen2025hac} have introduced entropy coding into 3DGS representations.
Building upon anchor-based designs~\cite{lu2024scaffold}, Chen \textit{et al.} proposed a Hash-grid Assisted Context (HAC)~\cite{chen2025hac}, while Wang \textit{et al.} introduced an anchor-level context\cite{wang2024contextgs} to enhance entropy modeling and further improve compression performance.

Despite recent progress in integrating such lossy-to-lossless compression strategies, current 3DGS compression methods still encounter two fundamental limitations:
1) For lossy compression, most existing methods~\cite{wang2024contextgs,chen2025hac} adopt a trivial quantization strategy that coarsely quantizes 3DGS data, while overlooking fine-grained (\textit{i.e.}, element-wise) quantization. This limits the ability to precisely control the bit-rate at the element level, thereby hindering optimal rate-distortion performance.
%
2) For lossless compression, most existing methods~\cite{chen2025hac,wang2024contextgs} adopt a conditional entropy model similar to those used in Neural Image Compression (NIC)~\cite{minnen2018joint,balle2018variational,cheng2020learned}, relying on a hyperprior to estimate the latent distribution. However, the design and expressiveness of the hyperprior remain limited. For instance, Chen \textit{et al}.~\cite{chen2025hac} directly adopted a shallow two-layer MLP to generate the hyperprior, which often fails to capture the full contextual dependencies needed for effective entropy modeling.


In this work, we propose a novel strategy, termed Mixture of Priors (MoP), to address both of the aforementioned limitations.
Inspired by the recent success of the Mixture of Experts (MoE) paradigm~\cite{shazeer2017moe,lepikhin2020moe1,fedus2022moe2,du2022moe3,xue2022moe4,zuo2022moe5} in foundation models, our proposed Mixture of Priors (MoP) strategy leverages multiple lightweight MLPs to extract diverse prior features and ensemble them to enhance both lossless and lossy compression performance. Specifically, each MLP is initialized with distinct parameters to promote the learning of diverse and specialized prior features, thereby improving the generalizability of the overall MoP representation. To ensure efficiency, all MLPs are designed to be lightweight, minimizing storage overhead. A learnable gating network dynamically assigns aggregation weights to each MLP output, enabling adaptive fusion of prior features into the final MoP feature.
%
The resulting MoP feature will serves as the hyperprior for conditional entropy coding, enabling more accurate distribution estimation and enhancing lossless compression performance.

Moreover, we leverage the produced MoP feature as guidance to implement a Coarse-to-Fine Quantization (C2FQ) mechanism, enabling element-wise quantization and improving lossy compression with more optimal rate-distortion performance. 
Specifically, we start with a predefined quantization step size, which is first refined into a quantization value and then adaptively expanded into a quantization vector under the guidance of the MoP feature. This vector is further scaled by the element-wise gradients of the 3DGS attributes to construct a quantization matrix, which is then used to perform element-wise quantization across all 3DGS attributes. This design enables fine-grained rate-storage adjustment at the element level. Notably, by leveraging the gradient information for this expansion rather than relying on auxiliary networks, we effectively avoid the memory overhead introduced by extra network parameters.


Extensive experimental results demonstrate that our proposed compression framework, with proposed MoP and C2FQ strategies, achieves state-of-the-art performance on various benchmark datasets, including Mip-NeRF360~\cite{barron2022mip}, BungeeNeRF~\cite{xiangli2022bungeenerf}, DeepBlending~\cite{hedman2018deep}, and Tank\&Temples~\cite{knapitsch2017tanks}. The main contributions of this work are summarized below:
\begin{itemize}
    \item We propose a novel Mixture of Priors strategy for 3DGS data compression, which employs multiple lightweight MLPs to generate diverse prior features and ensemble them into a unified MoP feature. This feature is used for both entropy coding and guiding quantization, thereby improving the performance of both lossy and lossless compression.

    \item Guided by the MoP features and element-wise gradients, we further propose a Coarse-to-Fine Quantization strategy that adaptively expands a predefined quantization step into an element-wise quantization matrix, enabling precise rate– storage adjustment for each individual element within the 3DGS attributes.
    

    \item We conduct comprehensive experiments on the Mip-NeRF360, BungeeNeRF, DeepBlending, and Tanks\&Temples benchmarks to demonstrate the effectiveness of our 3DGS compression framework. Equipped with the proposed MoP and C2FQ strategies, our method achieves state-of-the-art performance across these datasets.

\end{itemize}

\section{Related Work}
\subsection{3DGS Data Compression}
3DGS encodes 3D scenes using learnable geometric and appearance attributes represented as 3D Gaussian distributions. This approach delivers high-fidelity scene representation while supporting fast training and real-time rendering, contributing to its widespread adoption. However, the substantial number of Gaussians and their associated parameters introduces significant storage overhead, underscoring the need for effective 3DGS compression strategies.

Early methods primarily focused on reducing model complexity by refining Gaussian parameters. For example, vector quantization techniques grouped parameters into pre-defined codebooks~\cite{fan2023lightgaussian,lee2024compact,navaneet2023compact3d,niedermayr2024compressed, chen20254dgs}, while other approaches employed direct pruning to discard redundant components~\cite{fan2023lightgaussian,lee2024compact}.
More recent efforts~\cite{lu2024scaffold,morgenstern2023compact,chen2025hac,wang2024contextgs} have explored structural relationships to improve compression efficiency. Scaffold-GS~\cite{lu2024scaffold}, for instance, introduces anchor-centered features to represent scene content compactly. 
However, the above-mentioned methods do not use the entropy coding strategy to improve compression efficiency.
Based on Scaffold-GS, HAC~\cite{chen2025hac} first explores the entropy coding in 3DGS compression by utilizing a hash-grid structure to model spatial coherence, whereas ContextGS~\cite{wang2024contextgs} incorporates anchor-level contextual information as hyperprior information to achieve efficient entropy coding on 3DGS.



Despite their impressive performance, existing compression methods still exhibit several limitations. First, employing a single network with limited parameters inadequately extracts the prior features, leading to inefficient prediction of data distributions. Second, current methods lack fine-grained exploration of quantization steps, restricting their overall flexibility.


Therefore, we propose the MoP strategy and the C2FQ strategy to address the aforementioned limitations. Specifically, our MoP strategy employs multiple lightweight MLPs to extract different prior features for diversity improvement, enabling efficient prediction of data distributions and providing guidance for further quantization. Moreover, our C2FQ module leverages MoP features and element-wise gradients to achieve both large-scale and fine-grained adjustment of quantization steps, significantly enhancing entropy coding performance.



\begin{figure*}[t!]
    \centering
    \includegraphics[width=\linewidth]{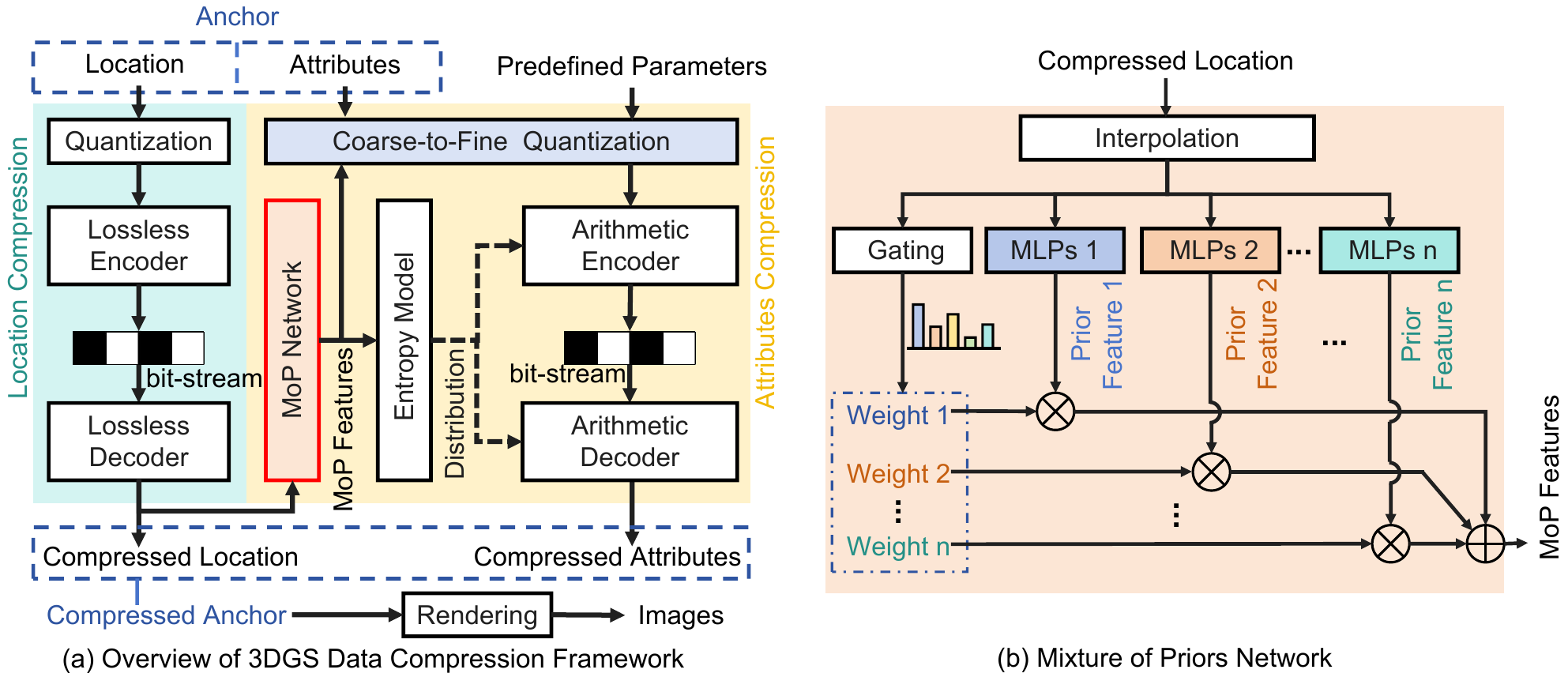}
    \caption{(a) The overview of our 3DGS compression framework, which integrates the proposed Mixture-of-Priors (MoP) network, the Coarse-to-Fine Quantization (C2FQ) module, and other standard 3DGS data compression components. (b) Details of the proposed MoP Network. It begins by applying a standard interpolation operation as in~\cite{chen2025hac} to extract features from the compressed location. These features are then used by a gating network and several lightweight MLPs to generate gating weights and diverse prior features. The prior features are subsequently aggregated into a unified MoP feature through a weighted summation based on the corresponding gating weight.}
    \label{fig:main_pipeline}
\end{figure*}

\begin{figure}[t!]
    \centering
    \includegraphics[width=\linewidth]{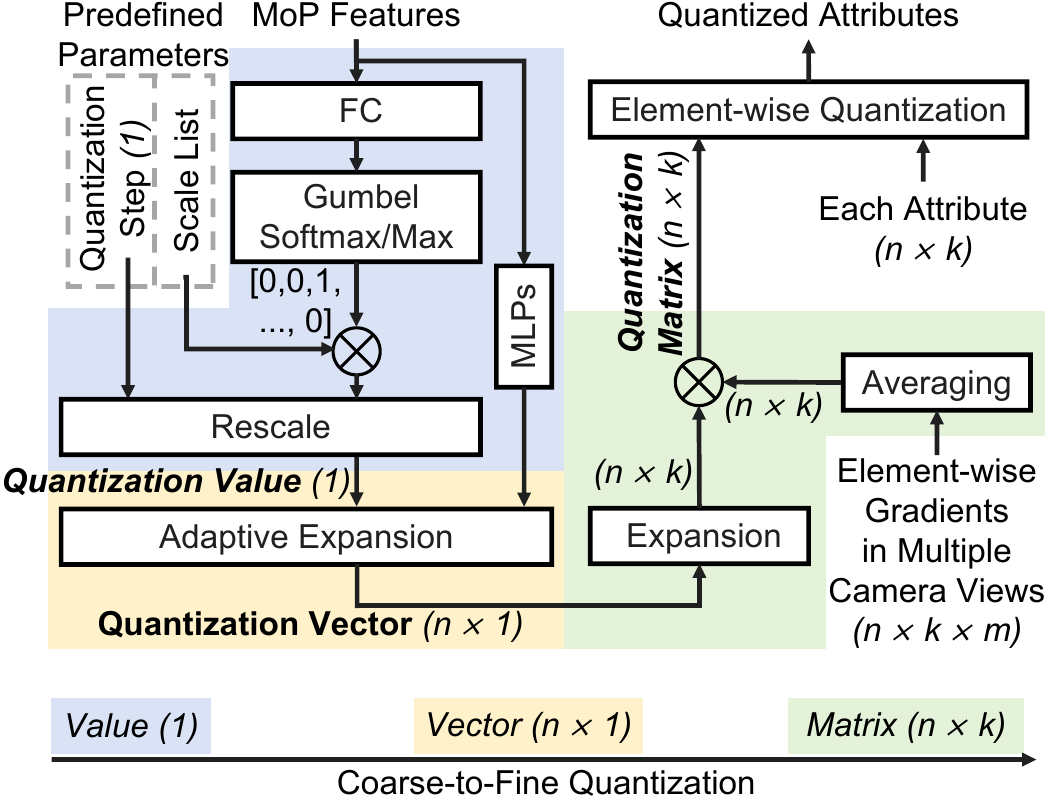}
    \caption{Details of our Coarse-to-Fine Quantization module. This module first rescales the predefined quantization step into a quantization value using a scale, which is selected from the scale list utilizing a Gumbel-Softmax/Max strategy.
    Further leveraging the MoP features, the adaptive expansion module expands this step value into a quantization vector. 
    Subsequently, the quantization vector is extended into a quantization matrix by multiplying it with aggregated element-wise gradients from multiple camera views. Finally, each attribute is quantized in an element-wise manner using the quantization matrix. The notations $(1)$, $(n\times 1)$, $(n\times k)$, and $(n\times k \times m)$ indicate the dimensions of the corresponding variables, where $n$ is the number of anchors, $k$ is the number of element in each anchor attribute, and $m$ is the number of camera views.}
    \label{fig:c2f}
\end{figure}

\subsection{Mixture of Experts}

The MoE framework~\cite{jacobs1991moe,shazeer2017moe} has been widely adopted due to its capability to extract diverse and rich data features from multiple expert networks~\cite{lepikhin2020moe1,fedus2022moe2,du2022moe3,xue2022moe4,zuo2022moe5}. However, directly applying traditional MoE models to 3DGS data compression still poses several limitations. Primarily, the large number of parameters typically used by MoE experts introduces prohibitive storage overhead for 3DGS compression tasks. To address this, we introduce multiple lightweight MLPs to extract rich features. This lightweight design significantly reduces the storage consumption associated with network parameters.
Additionally, given the limited feature extraction capability of each individual lightweight MLP, we employ a soft-gating mechanism to aggregate the outputs from all expert networks, thereby ensuring efficient utilization of each MLP.

\subsection{Quantization in Compression}

In traditional image and video compression algorithms (\textit{e.g.}, H.266~\cite{VVC}, and H.265~\cite{sullivan2012x265}), quantization steps are typically adjusted based on the characteristics of the coding elements, enabling more efficient quantization and improved compression performance. Similar ideas have increasingly been adopted in many compression methods~\cite{choi2020task,lee2022selective,luo2024super,10222717,xiang2018novel,fu2024weconvene,lei2023quantization,han2024cra5,liu2023icmh,chen2022exploiting,chen2024group,chen2023neural,hu2020improving,Chen_2022_CVPR,liu2025efficient,10219641,liu2024towards}, demonstrating the effectiveness of adaptive quantization in modern compression frameworks.

Although some 3DGS compression methods, such as HAC~\cite{chen2025hac}, attempt to adjust quantization steps, their adjustment range remains limited, and they fail to support element-level quantization. This restricts the overall performance of the compression models. To address this limitation, we propose a Coarse-to-Fine Quantization method. Leveraging MoP features, our approach enables a quantization vector. Furthermore, by accumulating the gradient of each element across multiple camera views and using it as weights, we refine the quantization vector to the quantization matrix. This improves the ability to precisely adjust the bit-rate at the element level, thereby enhancing the compression performance.

\section{Methodology}
\subsection{Preliminaries}
\textbf{3D Gaussian Splatting}~\cite{kerbl20233d} represents a 3D scene as a collection of Gaussians. Each Gaussian is defined by a location $\boldsymbol{\mu}$, a 3D covariance matrix $\bm{\mathit{\Sigma}}$, an opacity term $\alpha$ and a view-dependent color $\boldsymbol{c}$. Every Gaussian can be represented as: 
$ 
G(\boldsymbol{x}) = e^{-\frac{1}{2}(\boldsymbol{x} - \boldsymbol{\mu})^\mathrm{T} \bm{\mathit{\Sigma}}^{-1} (\boldsymbol{x} - \boldsymbol{\mu})}, 
$
where $\boldsymbol{x}$ denotes the coordinates of a 3D point. The covariance matrix $\bm{\mathit{\Sigma}}$ is further decomposed as $\bm{\mathit{\Sigma}} = \boldsymbol{R} \boldsymbol{S} \boldsymbol{S}^{\mathrm{T}} \boldsymbol{R}^{\mathrm{T}}$, with $\boldsymbol{R}$ and $\boldsymbol{S}$ corresponding to rotation and scaling components, respectively. When rendering 3DGS to images, these 3D Gaussians are splatted onto the given 2D plane, where pixel colors are determined by aggregating the splatted contributions using the opacity $\alpha$ and color $\boldsymbol{c}$.

\textbf{Anchor} is introduced by Scaffold-GS~\cite{lu2024scaffold}, which is a compact representation for 3DGS data. Specifically, each anchor is defined by a 3D location $\bm{x}^{a} \in \mathbb{R}^{3}$ and a set of associated attributes $\bm{\mathcal{A}} = {\bm{f}^{a} \in \mathbb{R}^{D^{a}}, \bm{l} \in \mathbb{R}^{6}, \bm{o} \in \mathbb{R}^{3K}}$, where $\bm{f}^{a}$ denotes the anchor’s local feature vector, $\bm{l}$ represents scaling factors, and $\bm{o}$ specifies positional offsets.
During rendering, the feature $\bm{f}^{a}$ is passed through MLPs to predict the properties of the associated Gaussians. The positions of these Gaussians are computed by applying the offsets $\bm{o}$ to the anchor location $\bm{x}^{a}$, while the scaling vector $\bm{l}$ modulates their spatial extent and shape.

\textbf{Quantization} is a technique that maps continuous-valued signals to a finite set of discrete levels, enabling lossy compression by reducing data precision.
A predefined quantization step value determines the degree of data discretization, directly impacting storage cost. In this work, we adopt an element-wise quantization strategy that utilizes this step value to quantize each element.

\textbf{Entropy coding} is a fundamental technique used to achieve lossless compression by efficiently encoding data based on its statistical distribution.
It relies on the probability distribution of the quantized item $\bm{\mathcal{\hat{A}}}$ for efficient encoding. Since the true distribution $q(\bm{\mathcal{\hat{A}}})$ is generally inaccessible, it is commonly approximated by an estimated distribution $p(\bm{\mathcal{\hat{A}}})$~\cite{balle2018variational,minnen2018joint,li2023dcvcdc,li2024dcvcfm,dvc,fvc,li2021dcvc}. According to \textit{Information Theory}~\cite{cover1999elements}, the expected number of bits required to encode $\bm{\mathcal{\hat{A}}}$ using entropy coding is given by the cross-entropy, defined as $H(q,p)=\mathbb{E}_{\bm{\mathcal{\hat{A}}} \sim q}[-\log(p(\bm{\mathcal{\hat{A}}}))]$. This cross-entropy serves as a lower bound on the achievable storage. Therefore, improving the accuracy of the approximation $p(\bm{\mathcal{\hat{A}}})$ reduces $H(q,p)$ and leads to lower storage cost.
In this work, we model the conditional distribution $p(\bm{\mathcal{\hat{A}}} | \cdot)$ of the quantized attribute $\mathcal{\hat{A}}$ and apply arithmetic coding as the cross-entropy coding algorithm.



\subsection{The Overview of our 3DGS Data Compression Framework}

As a compact representation of 3DGS, the anchor has recently gained significant attention in 3DGS compression and has demonstrated impressive performance. Motivated by the advantages of anchors, we propose a novel anchor-based 3DGS compression framework, as illustrated in Figure~\ref{fig:main_pipeline} (a). An anchor primarily consists of two parts: location $\bm{x}^{a}$ and attributes $\bm{\mathcal{A}}$. Accordingly, our framework is designed with two separate branches—a location compression branch and an attributes compression branch—which are responsible for compressing the anchor's location and attributes, respectively.
Specifically, the framework first compresses the anchor locations to obtain compact location representations. These compressed locations are subsequently utilized to guide the compression of anchor attributes. Finally, the compressed location and attributes are combined to form the compressed anchors, which are passed to a rendering module to reconstruct the 3D Scene and render the final images. The details of the entire network are described as follows:

\textbf{Location Compression.} Following the standard practice in current 3DGS compression methods~\cite{chen2025hac,wang2024contextgs,lee2024compact,papantonakis2024reducing}, we construct a location compression pipeline accordingly.
Specifically, we first quantize the anchor locations $\bm{x}^a$ from 32-bit precision to 16-bit. The quantized locations $\bm{\hat{x}}^a$ are then losslessly encoded into a bit-stream. Subsequently, the bit-stream is losslessly decoded to obtain the compressed locations $\bm{\bar{x}}^a$. 


\textbf{MoP Network.} 
The compressed location is further fed into our proposed MoP Network to generate MoP feature $\bm{\mathcal{G}}$, which serves both as guidance for quantization and as hyperprior information for enabling lossy and lossless compression.
More details of the MoP network are provided in Section~\ref{sec:moe}.

\textbf{Lossy Attribute Compression.} 
We achieve lossy compression of the anchor attributes through quantization, guided by the MoP features generated by the MoP Network. Specifically, the Coarse-to-Fine Quantization procedure expands a predefined quantization step value into a quantization matrix, enabling element-wise quantization of each individual element within the anchor. More details are provided in Section~\ref{sec:c2f}.

\textbf{Lossless Attribute Compression.} 
To perform lossless compression of the quantized attribute $\bm{\mathcal{\hat{A}}}$, we directly use the MoP feature $\bm{\mathcal{G}}$ generated by the MoP Network as hyperprior information to predict the distribution $p(\bm{\mathcal{\hat{A}}}|\bm{\mathcal{G}})$ with a conditional entropy model, following~\cite{chen2025hac}.
Based on such estimated distribution $p(\bm{\mathcal{\hat{A}}}|\bm{\mathcal{G}})$,
an arithmetic encoder losslessly encodes the quantized attributes $\bm{\mathcal{\hat{A}}}$ into a bit-stream. During the decoding phase, the bit-stream is losslessly decoded by an arithmetic decoder using the same distribution to reconstruct the quantized attributes $\bm{\mathcal{\bar{A}}}$.

\textbf{Rendering.} The compressed location $\bm{\bar{x}}^a$ and attributes $\bm{\mathcal{\bar{A}}}$ together form the compressed anchors. We follow the standard anchor-based rendering pipeline~\cite{lu2024scaffold} to generate images from these anchors. Specifically, the anchors are first reconstructed into 3DGS. Given the camera viewpoint, the 3DGS is then projected onto the 2D image plane, and pixel values are rendered using $\alpha$-composited blending, as described in~\cite{lu2024scaffold,chen2025hac,wang2024contextgs}.

\subsection{Mixture of Priors}
\label{sec:moe}

The MoP Network is utilized to extract MoP features from the compressed locations for both lossy and lossless compression. The architecture of the MoP Network is illustrated in Figure~\ref{fig:main_pipeline} (b). The MoP network first interpolates the compressed location following the approach in~\cite{chen2025hac}, and the resulting interploated location information is then used to construct the MoP features.




Different with previous 3DGS compression networks~\cite{chen2025hac,wang2024contextgs} that employ a single MLP to exploit the hyperprior information
Inspired by the MoE paradigms~\cite{shazeer2017moe,lepikhin2020moe1,fedus2022moe2,du2022moe3,xue2022moe4,zuo2022moe5}, we design multiple lightweight MLPs to extract diverse prior features $\mathbf{p_1}, \mathbf{p_2}, ..., \mathbf{p_n}$ from the interpolated location information. 
Meanwhile, a gating network produces a set of weights $[w_1, w_2, ..., w_n]$ corresponding to each prior feature. These weights are then applied to their respective prior features through element-wise multiplication, resulting in weighted prior features. Finally, all weighted prior features are summed to obtain the MoP feature as follows:
\begin{equation}
\bm{\mathcal{G}} = \sum_{i = 1}^{n} w_{i} \times \mathbf{p}_{i}
\end{equation}
where $\bm{\mathcal{G}}$ represents the MoP features explored by our MoP Network, $\mathbf{p}_i$ denotes the output feature of the $i$-th MLPs, and $w_i$ is the corresponding weight generated by the gating network.

In this process, instead of using a single MLP to explore the priors as in previous 3DGS compression methods~\cite{chen2025hac,wang2024contextgs}, we design multiple MLPs to extract diverse prior features. To ensure feature diversity, we adopt different parameter initialization strategies for each MLP, encouraging them to learn distinct parameters during training and thus generate diverse prior features.
While considering that network parameters must be stored with the compressed data and thus contribute to storage overhead, we adopt a lightweight design for the gating network and each MLP, avoiding the negative impact of excessive parameters on compression performance. 
Owing to the aforementioned advantages, the MoP features extracted by our MoP network play a crucial role in both lossless and lossy compression. Specifically, for lossless compression, the diverse MoP features provide richer contextual information for the entropy model to predict more accurate data distributions, thereby improving compression efficiency. For lossy compression, the MoP features guide the quantization process, enabling precise adjustment of quantization steps, as described in Section~\ref{sec:c2f}.

\subsection{Coarse-to-Fine Quantization}
\label{sec:c2f}



Quantization is an essential component in lossy compression, as it directly affects the storage costs when encoding the data. To precisely adjust the storage costs of the data, element-wise quantization strategies have been explored in both traditional~\cite{VVC,sullivan2012x265} and deep learning-based~\cite{choi2020task,lee2022selective,luo2024super,10222717,xiang2018novel,fu2024weconvene,lei2023quantization} image and video compression. However, such techniques remain largely unexplored in 3DGS compression. To address this gap, we propose a Coarse-to-Fine Quantization strategy guided by our MoP features that adjusts the quantization step in an element-wise way.




The details of our Coarse-to-Fine Quantization module are shown in Figure~\ref{fig:c2f}.
In the initial adjustment stage, our goal is to obtain a coarse quantization step from a large scale, which serves as the basis for subsequent fine-gained quantization.
To this end, a scale list is provided to the quantization module.
Our strategy employs the MoP feature to compute the probability of each scale through one fully connected (FC) layer. The scale with the highest probability is then selected via a max operation.
However, since the max operation is non-differentiable, it prevents end-to-end optimization of the entire network through backpropagation. To address this issue, we adopt the Gumbel-Softmax strategy~\cite{gumbel} during training to approximate the selection process. As the Gumbel-Softmax module is differentiable, it enables end-to-end optimization of the network.
After the scale $s$ selected by the Gubmel-Softmax/Max strategy, our C2FQ rescales the predefined quantization step $Q_0$ to the quantization value $Q_1$ as follows: $Q_1 = Q_0 \times s$.

Subsequently, the C2FQ strategy further leverages the MoP feature $\bm{\mathcal{G}}$ to adaptively expand the quantization value $Q_1$ into a quantization vector $\bm{Q}_2$. This expansion is defined as $\bm{Q}_2 = Q_1 \times (1+\text{Tanh}(f_{\phi}(\bm{\mathcal{G}})))$, where $f_{\phi}$ is the MLPs. 

To further expand the quantization vector $\bm{Q}_2$ to element-wise, the most straightforward approach is to introduce a neural network that expands the $\bm{Q}_2$ to a 
$(n\times k)$-dimensional matrix, where $n$ is the number of anchors and $k$ is the number of elements in each anchor attribute. However, for large values of 
$k$, this introduces a significant number of network parameters, resulting in substantial storage overhead and negatively affecting compression performance.
To avoid this issue, we exploit the relationship between an element's gradient and its contribution to the loss: elements with larger gradients have a greater impact on the loss and, consequently, on the final compression performance, while those with smaller gradients contribute less. Based on this insight, we propose a network-free strategy to assign the element-wise quantization matrix $\bm{Q}_3$ using gradient-based weighting.
Specifically, we first expand the quantization vector $\bm{Q}_2$ of shape 
$(n\times 1)$ to shape $(n\times k)$ by duplicating it 
$k$ times. 
Next, we compute the average gradient across all camera views for each element. These averaged gradients are then used as weights and multiplied element-wise with the expanded quantization steps to generate a quantization matrix $\bm{Q}_4$ for each elements.

Finally, the attribute values $\bm{\mathcal{A}}$ are quantized using the computed fine-grained quantization steps $\bm{Q}_4$, following $\bm{\mathcal{\hat{A}}} = \mathrm{Round}(\bm{\mathcal{A}} \times \bm{Q}_4)/\bm{Q}_4$. Through this process, we achieve element-level, fine-grained quantization of the attributes, leading to precisely adjusting the storage for each element in lossy compression.

\begin{table*}[h]
\centering
\setlength{\tabcolsep}{2pt}
\renewcommand{\arraystretch}{1.3}

\caption{A comparative analysis of the newly proposed method with other 3DGS data compression methods. Two distinct sets of results are reported for our method, reflecting varying trade-offs between size and fidelity. The best and second-best results are marked in \colorbox{pink}{\textcolor{black}{red}} and \colorbox{yellow!50}{\textcolor{black}{yellow}} cells, respectively. Size measurements are provided in megabytes (MB).}
\label{tab:main_results}

\begin{tabular}{l|cccc|cccc|cccc|cccc}
\hline
\rowcolor[HTML]{FFFFFF} 
\textbf{Datasets}  & \multicolumn{4}{c|}{\textbf{Mip-NeRF360~\cite{barron2022mip}}} & \multicolumn{4}{c|}{\textbf{BungeeNeRF~\cite{xiangli2022bungeenerf}}} & \multicolumn{4}{c|}{\textbf{DeepBlending~\cite{hedman2018deep}}} & \multicolumn{4}{c}{\textbf{Tank\&Temples~\cite{knapitsch2017tanks}}} \\ \hline
\rowcolor[HTML]{FFFFFF} 
\textbf{Methods} & \textbf{psnr$\uparrow$} & \textbf{ssim$\uparrow$} & \textbf{lpips$\downarrow$} & \textbf{size$\downarrow$}  & \textbf{psnr$\uparrow$} & \textbf{ssim$\uparrow$} & \textbf{lpips$\downarrow$} & \textbf{size$\downarrow$}  & \textbf{psnr$\uparrow$} & \textbf{ssim$\uparrow$} & \textbf{lpips$\downarrow$} & \textbf{size$\downarrow$}  & \textbf{psnr$\uparrow$} & \textbf{ssim$\uparrow$} & \textbf{lpips$\downarrow$} & \textbf{size$\downarrow$} \\ \hline
3DGS~\cite{kerbl20233d} & 27.49 & \cellcolor{pink}{0.813} & \cellcolor{pink}{0.222} & 744.7 & 24.87 & 0.841 & \cellcolor{yellow!50}{0.205} & 1616 & 29.42 & 0.899 & \cellcolor{pink}{0.247} & 663.9 & 23.69 & 0.844 & 0.178 & 431.0 \\ 
Scaffold-GS~\cite{lu2024scaffold} & 27.50 & 0.806 & 0.252 & 253.9 & 26.62 & 0.865 & 0.241 & 183.0 & 30.21 & 0.906 & 0.254 & 66.00 & 23.96 & 0.853 & 0.177 & 86.50 \\ \hline
EAGLES~\cite{girish2023eagles} & 27.15 & 0.808 & 0.238 & 68.89 & 25.24 & 0.843 & 0.221 & 117.1 & 29.91 & \cellcolor{yellow!50}{0.910} & \cellcolor{yellow!50}{0.250} & 62.00 & 23.41 & 0.840 & 0.200 & 34.00 \\ 
LightGaussian~\cite{fan2023lightgaussian} & 27.00 & 0.799 & 0.249 & 44.54 & 24.52 & 0.825 & 0.255 & 87.28 & 27.01 & 0.872 & 0.308 & 33.94 & 22.83 & 0.822 & 0.242 & 22.43 \\ 
Compact3DGS~\cite{lee2024compact} & 27.08 & 0.798 & 0.247 & 48.80 & 23.36 & 0.788 & 0.251 & 82.60 & 29.79 & 0.901 & 0.258 & 43.21 & 23.32 & 0.831 & 0.201 & 39.43 \\ 
Compressed3D~\cite{navaneet2023compact3d} & 26.98 & 0.801 & 0.238 & 28.80 & 24.13 & 0.802 & 0.245 & 55.79 & 29.38 & 0.898 & 0.253 & 25.30 & 23.32 & 0.832 & 0.194 & 17.28 \\ 
Morgen. \textit{et al.}~\cite{morgenstern2023compact} & 26.01 & 0.772 & 0.259 & 23.90 & 22.43 & 0.708 & 0.339 & 48.25 & 28.92 & 0.891 & 0.276 & 8.40 & 22.78 & 0.817 & 0.211 & 13.05 \\ 
Navaneet \textit{et al.}~\cite{navaneet2023compact3d} & 27.16 & 0.808 & 0.228 & 50.30 & 24.63 & 0.823 & 0.239 & 104.3 & 29.75 & 0.903 & \cellcolor{pink}{0.247} & 42.77 & 23.47 & 0.840 & 0.188 & 27.97 \\ 
Reduced3DGS~\cite{papantonakis2024reducing} & 27.19 & 0.807 & 0.230 & 29.54 & 24.57 & 0.812 & 0.228 & 65.39 & 29.63 & 0.902 & 0.249 & 18.00 & 23.57 & 0.840 & 0.188 & 14.00 \\
RDOGaussian~\cite{wang2024end} & 27.05 & 0.802 & 0.239 & 23.46 & 23.37 & 0.762 & 0.286 & 39.06 & 29.63 & 0.902 & 0.252 & 18.00 & 23.34 & 0.835 & 0.195 & 12.03 \\
MesonGS~\cite{xie2024mesongs} & 26.99 & 0.796 & 0.247 & 27.16 & 23.06 & 0.771 & 0.235 & 63.11 & 29.51 & 0.901 & 0.251 & 24.76 & 23.32 & 0.837 & 0.193 & 16.99 \\
CompGS~\cite{liu2024compgs} & 27.26 & 0.803 & 0.239 & \cellcolor{yellow!50}{16.50} & - & - & - & - & 29.69 & 0.901 & 0.279 & 8.77 & 23.70 & 0.837 & 0.208 & \cellcolor{yellow!50}{9.60} \\
HAC~\cite{chen2025hac}  &  \cellcolor{yellow!50}{27.77} & \cellcolor{yellow!50}{0.811} & 0.230 & 21.87 & 27.08 & 0.872 & 0.209 & 29.72 & 30.34 & 0.906 & 0.258 &  6.35 & \cellcolor{yellow!50}{24.40} & 0.853 & 0.177 & 11.24 \\ 

Context-GS~\cite{wang2024contextgs} & 27.72 & \cellcolor{yellow!50}{0.811} & 0.231 & 21.58 & 27.15 & \cellcolor{yellow!50}{0.875} & \cellcolor{yellow!50}{0.205} & \cellcolor{yellow!50}{21.80} & \cellcolor{yellow!50}{30.39} & 0.909 & 0.258 & 6.60 & 24.29 & 0.855 & 0.176 & 11.80 \\ \hline

Ours (low-rate) & 27.68 & 0.808 & 0.234 & \cellcolor{pink}{15.64} & \cellcolor{yellow!50}{27.26} & \cellcolor{yellow!50}{0.875} & 0.207 & \cellcolor{pink}{20.83} & 30.20 & 0.908 & 0.260 & \cellcolor{pink}{4.07} & 24.21 & \cellcolor{yellow!50}{0.861} & \cellcolor{yellow!50}{0.163} & \cellcolor{pink}{8.98} \\ 

Ours (high-rate) & \cellcolor{pink}{27.89} & \cellcolor{yellow!50}{0.811} & \cellcolor{yellow!50}{0.227} & 21.89 & \cellcolor{pink}{27.63} & \cellcolor{pink}{0.893} & \cellcolor{pink}{0.172} & 27.56 & \cellcolor{pink}{30.45} & \cellcolor{pink}{0.912} & \cellcolor{yellow!50}{0.250} & \cellcolor{yellow!50}{5.65} & \cellcolor{pink}{24.43} & \cellcolor{pink}{0.865} & \cellcolor{pink}{0.158} & 11.33 \\ \hline
\end{tabular}
\end{table*}

\subsection{Optimization}
The total loss function used to optimize the proposed 3DGS compression framework is formulated as follows:
\begin{equation}
    \mathcal{L} = \mathcal{L}_{\mathrm{Rendering}} + \lambda\mathcal{L}_{\mathrm{anchor}},
\end{equation}
where $\mathcal{L}_{\mathrm{Rendering}}$ is the rendering loss defined in Scaffold-GS~\cite{lu2024scaffold}, and $\mathcal{L}_{\mathrm{anchor}}$ denotes the estimated storage cost for anchor as in HAC~\cite{chen2025hac} and Context-GS~\cite{wang2024contextgs}. Specifically, $\mathcal{L}_{\mathrm{anchor}}$ mainly reflects the estimated storage cost of anchor attributes derived from entropy coding. $\lambda$ is a hyperparameter that balances the different loss components.

\section{Experiments}

\begin{figure*}[htbp]

    \begin{center}
        {
         \label{fig:sub_fig3}\includegraphics[width=0.32\textwidth]{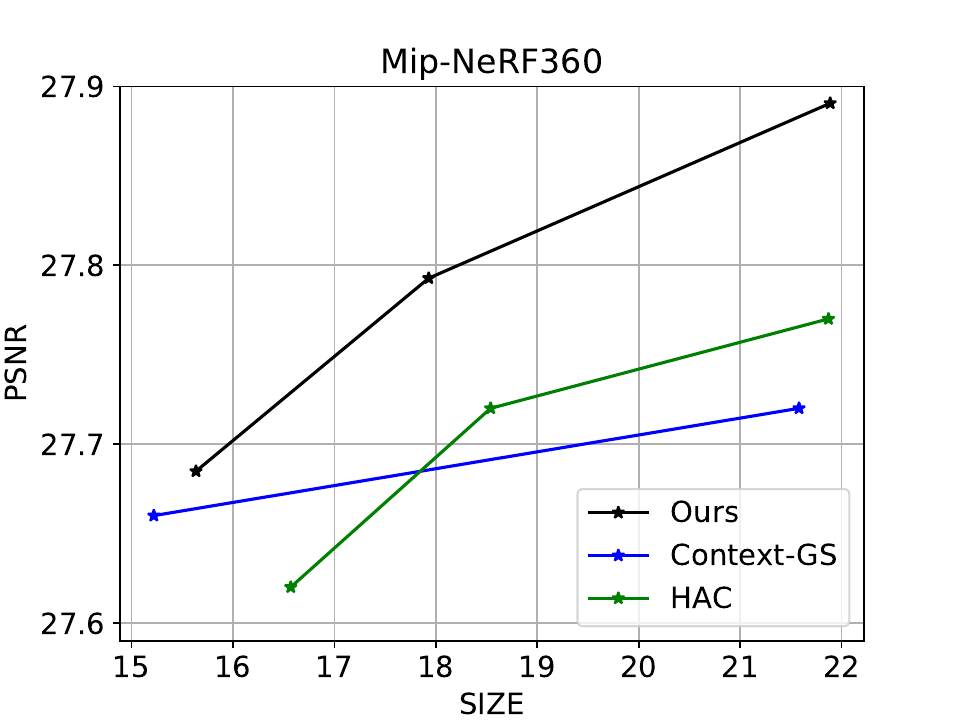}
        }
        {
         \label{fig:sub_fig2}\includegraphics[width=0.32\textwidth]{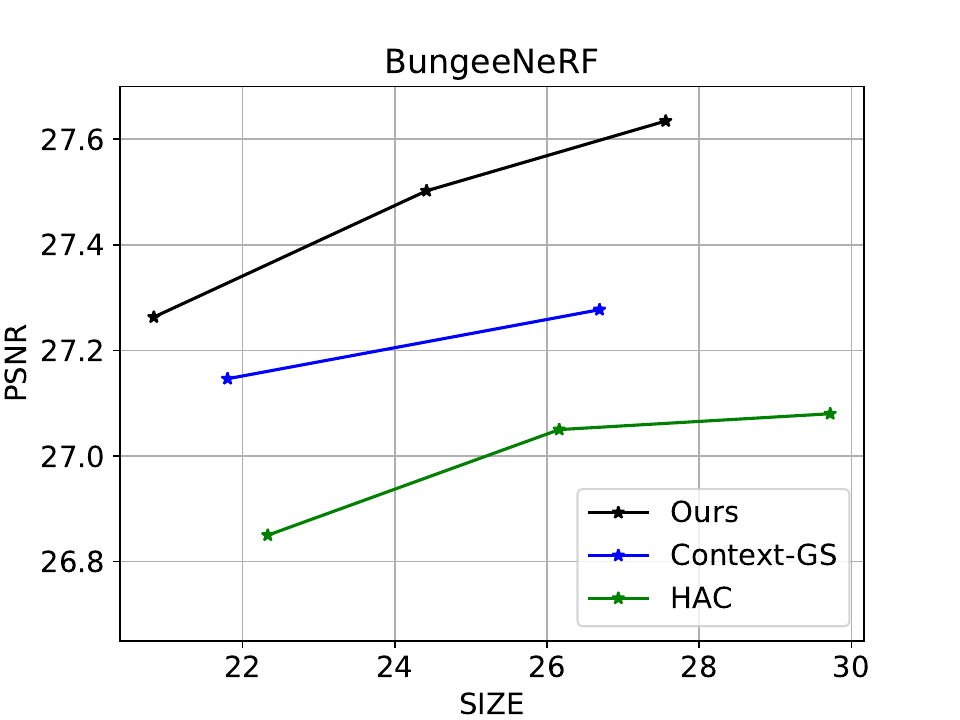}
        }
        {
         \label{fig:sub_fig1}\includegraphics[width=0.32\textwidth]{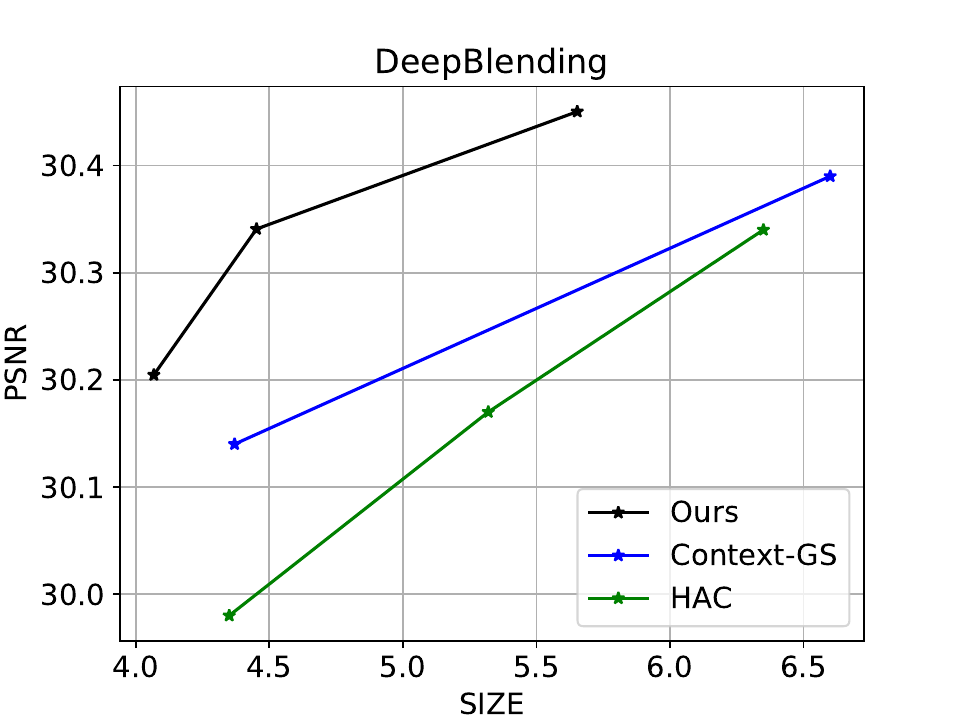}
        }
    \end{center}
    \caption{The Rate-Distortion (RD) curves on three benchmarks, including Mip-NeRF360, BungeeNeRF, and DeepBlending.}
    \label{fig:RD}
\end{figure*}

\subsection{Datasets}
We performed extensive evaluations across multiple datasets, including the four large-scale real-world benchmarks: Mip-NeRF360~\cite{barron2022mip}, BungeeNeRF~\cite{xiangli2022bungeenerf}, DeepBlending~\cite{hedman2018deep}, and Tanks\&Temples~\cite{knapitsch2017tanks}.





\subsection{Experiment Details}
\textbf{Baseline methods.} We compare our newly proposed 3DGS compression method against a range of existing 3DGS compression approaches. Several prior works~\cite{fan2023lightgaussian,lee2024compact,navaneet2023compact3d,niedermayr2024compressed} focus on reducing model size via parameter pruning or vector quantization with codebooks. Other approaches~\cite{chen2025hac,wang2024contextgs} aim to enhance 3DGS data compression by incorporating entropy coding. Among them, HAC~\cite{chen2025hac} leverages hash grids, while ContextGS~\cite{wang2024contextgs} utilizes anchor-level context as a hyperprior to facilitate effective entropy coding of 3DGS data.

\textbf{Metrics.} We evaluate compression performance in terms of storage size, measured in megabytes (MB). To assess the visual quality of rendered images generated from the compressed 3DGS data, we employ three standard metrics: Peak Signal-to-Noise Ratio (PSNR), Structural Similarity Index (SSIM)\cite{wang2004ssim}, and Learned Perceptual Image Patch Similarity (LPIPS)\cite{zhang2018LPIPS}.


\textbf{Implementation Details.} Our method is implemented in PyTorch with CUDA acceleration, and all experiments are conducted on the machine with Intel Xeon CPU and an NVIDIA RTX 3090 GPU equipped with 24GB of memory. The model is optimized using the Adam optimizer~\cite{kingma2014adam} and trained for 60,000 iterations. To balance storage overhead and prior feature diversity, we empirically set the number of MLP modules in our MoP strategy to five. When computing element-wise gradients for quantization, we first compress the anchors under multiple camera views. At this stage, the element gradients fed into the Coarse-to-Fine Quantization module are initially set to 1, meaning that fine-grained quantization is not applied. Once the gradients are collected, we then feed the actual element-wise gradients into the quantization module to enable fine-grained quantization.


\begin{table}[t!]
\centering
\caption{Ablation study on the DeepBlending dataset. (1) \textbf{Ours}: the full version of our proposed 3DGS compression framework. (2) \textbf{Ours w/o C2FQ}: our method without the Coarse-to-Fine Quantization strategy. (3) \textbf{Ours w/o MoP}: our method without the MoP strategy. (4) \textbf{Ours w/o C2FQ \& MoP}: our method without both the Coarse-to-Fine Quantization strategy and the MoP strategy.}
\begin{tabular}{l|cccc}
    \hline
        & PSNR$\uparrow$ & SSIM$\uparrow$ & Size (MB)$\downarrow$ \\ \hline
         Ours & \textbf{30.45} & \textbf{0.912} & \textbf{5.65}  \\ 
         Ours w/o C2FQ & 30.39 & 0.911 & 5.78  \\ 
         Ours w/o MoP & 30.23 & 0.910 & 5.75 \\
         Ours w/o C2FQ \& MoP & 30.22 & 0.908 & 5.81 \\ \hline
    \end{tabular}
    \label{tab:ab}
\end{table}

\subsection{Experiment Results}
\begin{figure*}[t!]
    \centering
    \includegraphics[width=\linewidth]{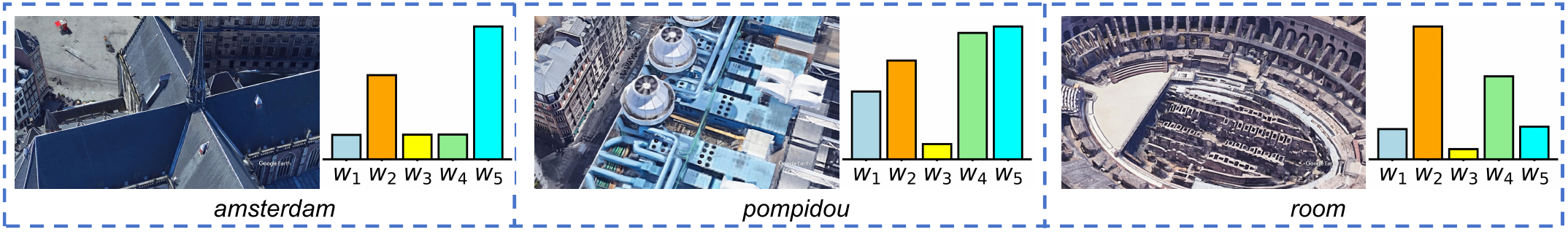}
    \caption{Visualization of ``amsterdam'', ``pompidou'', and ``room'' scenes from BungeeNeRF dataset, and the corresponding weights [$w_1$, $w_2$, $w_3$, $w_4$, $w_5$] for different prior features.}
    \label{fig:weights}
\end{figure*}

\begin{table}[t!]
    \centering
    \caption{BDBR (\%) results for HAC, Context-GS, and our newly proposed 3DGS Compression method across different datasets. Positive BDBR values indicate additional storage costs compared to our method.}
    \begin{tabular}{c|cc}
    \hline
        & Mip-NeRF360 & DeepBlending \\ \hline
         Context-GS~\cite{wang2024contextgs} & 22.50 & 31.80  \\ \hline
         HAC~\cite{chen2025hac} & 16.55 & 41.82  \\ \hline
    \end{tabular}
    \label{tab:bdbr}
\end{table}

We compare our method with existing 3DGS compression approaches on four benchmark datasets: Mip-NeRF360~\cite{barron2022mip}, BungeeNeRF~\cite{xiangli2022bungeenerf}, DeepBlending~\cite{hedman2018deep}, and Tanks\&Temples~\cite{knapitsch2017tanks}. The experimental results are presented in Table~\ref{tab:main_results}. Our method reduces storage by more than 97\% compared to the original 3DGS~\cite{kerbl20233d}, while achieving even higher rendering fidelity. Compared with Scaffold-GS~\cite{lu2024scaffold}, our approach achieves over 88\% storage savings and consistently delivers better reconstruction quality. These results demonstrate the effectiveness of our compression framework in significantly reducing the storage cost of 3DGS. Compared with other recent compression approaches~\cite{kerbl20233d,lu2024scaffold,girish2023eagles,fan2023lightgaussian,lee2024compact,morgenstern2023compact,navaneet2023compact3d,papantonakis2024reducing,wang2024end,xie2024mesongs,liu2024compgs,chen2025hac,wang2024contextgs}, our method consistently achieves the best performance.


Furthermore, to more clearly compare with the recent state-of-the-art methods—HAC~\cite{chen2025hac} and Context-GS~\cite{wang2024contextgs}—we present performance comparisons across different storage sizes, as shown in Figure~\ref{fig:RD}. 
To provide a more comprehensive evaluation, we also report the Bjøntegaard Delta Bit Rate (BDBR) results in Table~\ref{tab:bdbr}. These results show that, under the same PSNR setting, our method reduces the average storage cost by 31.80\% and 41.82\% compared to HAC and Context-GS, respectively, on the DeepBlending dataset. Overall, these experiments confirm that our method outperforms prior state-of-the-art approaches, demonstrating its effectiveness in 3DGS compression.

\subsection{Ablation Study and Analysis}

\textbf{Effectiveness of Different Components.} As shown in Table~\ref{tab:ab}, we take the DeepBlending~\cite{hedman2018deep} dataset as an example to evaluate the effectiveness of the key components in our proposed 3DGS compression framework. To assess the impact of the Coarse-to-Fine Quantization module, we remove it from our method, denoted as ``Ours w/o C2FQ''. Compared to the full framework, ``Ours w/o C2FQ'' results in a drop of 0.06 PSNR and 0.001 SSIM, along with an increase of 0.13MB storage cost. To assess the impact of the MoP strategy, we remove it from our method, denoted as ``Ours w/o MoP''. Compared to the full framework, ``Ours w/o MoP'' results in a drop of 0.22 PSNR and 0.002 SSIM, along with an increase of 0.10MB storage cost. Furthermore, we remove the MoP strategy from ``Ours w/o C2FQ'', resulting in the variant ``Ours w/o C2FQ \& MoP''. This leads to a further decrease of 0.05 PSNR and 0.005 SSIM, and increases the storage cost by an additional 0.57MB compared to ``Ours w/o C2FQ''. These results clearly demonstrate the effectiveness of both the Coarse-to-Fine Quantization module and the MoP Network in improving compression performance.

\begin{table}[t!]
    \centering
    \caption{Comparison of MLPs size and compression performance (\textit{i.e.}, PSNR, and total size) between HAC, Context-GS, and our ``Ours w/o C2FQ'' variant on the DeepBlending dataset. Here, ``Ours w/o C2FQ'' refers to our method without the Coarse-to-Fine Quantization module.}
    \begin{tabular}{c|ccc}
    \hline 
        \multirow{2}{*}{ }& \multirow{2}{*}{\makecell{Size (MB) of \\ MLPs $\downarrow$ }}& \multirow{2}{*}{PSNR$\uparrow$} & \multirow{2}{*}{\makecell{Total \\ Size (MB) $\downarrow$}} \\ 
        & & & \\ \hline
         Ours w/o C2FQ & 0.378 & \textbf{30.39} & \textbf{5.78} \\
         HAC~\cite{chen2025hac} & \textbf{0.157} & 30.34 & 6.35 \\
         Context-GS~\cite{wang2024contextgs} & 0.316 & \textbf{30.39} & 6.60 \\ \hline
    \end{tabular}
    \label{tab:mop}
\end{table}

\begin{table*}
    \centering
    \caption{The storage cost of each component and the rendering qualities of HAC and our newly proposed compression framework on the ``Garden'' scene of the Mip-NeRF360 dataset. ``Others'' means additional storage costs.}
    \begin{tabular}{c|ccccc|ccc} \hline
    \multirow{2}{*}{Methods}& \multicolumn{5}{c|}{Storage Costs (MB)$\downarrow$} & \multicolumn{3}{c}{Fidelity} \\ \cline{2-9}
         & Location & Attributes & MLPs & Others & Total & PSNR$\uparrow$ & SSIM$\uparrow$ & LPIPS$\downarrow$ \\ \hline

     Ours & \textbf{3.10} & \textbf{20.25} & 0.38 & \textbf{0.73} & \textbf{24.45} & \textbf{27.51} & \textbf{0.851} & \textbf{0.135}  \\ 
     HAC~\cite{chen2025hac} & 4.01 & 27.24 & \textbf{0.16} & 0.94 & 32.35 & 27.50 & \textbf{0.851} & 0.138  \\
         \hline
    \end{tabular}
    \label{tab:storage_cost}
\end{table*}

\textbf{Analysis of our MoP strategy.} Our MoP strategy employs lightweigth experts to extract diverse hyperprior features for distribution prediction, effectively balancing the trade-off between the storage cost of network parameters and compression performance. To isolate the contribution of the MoP strategy, we remove the Coarse-to-Fine quantization module from our complete framework and retain only the MoP network, denoted as ``Ours w/o C2FQ''. We compare ``Ours w/o C2FQ'' with HAC~\cite{chen2025hac} and Context-GS~\cite{wang2024contextgs}, both of which do not explicitly address the trade-off between model complexity and compression performance. The comparison is conducted on the DeepBlending~\cite{hedman2018deep} dataset, and the results are presented in Table~\ref{tab:mop}. 
The results indicate that although the use of multiple lightweight MLPs introduces a slight increase in parameter size, the diverse hyperprior features extracted by them significantly improve the performance of the compression model (\textit{i.e.}, higher PSNR and lower overall storage consumption).

Furthermore, we visualize the prior weights of different scenes from BungeeNeRF~\cite{xiangli2022bungeenerf} dataset, as shown in Figure~\ref{fig:weights}. The visualization results demonstrate that the gating network in our MoP network can adaptively adjust the weights for different prior features across scenes. This shows that the MoP strategy not only preserves the diversity of prior features but also emphasizes scene-relevant information, enabling the predicted distribution that is better suited to each specific scene. Consequently, the compression performance is further improved.

\begin{table}[t!]
    \centering
    \caption{Ablation study of different quantization stages on the Tank\&Temples dataset. (1) \textbf{Ours:} the full version of our proposed 3DGS compression method. (2) \textbf{Ours w/o QM}: Our method removes the quantization matrix. (3) \textbf{Ours w/o QM \& QV}: Our method removes both the quantization matrix and vector.}
    \begin{tabular}{l|cc}
    \hline
            & PSNR $\uparrow$ & Size (MB) $\downarrow$ \\ \hline
        Ours & \textbf{24.37} & \textbf{10.35} \\
        Ours w/o QM & 24.36 & 10.50 \\ 
        Ours w/o QM \& QV & 24.23 & 10.73 \\ \hline
    \end{tabular}
    \label{tab:c2f}
\end{table}

\textbf{Analysis of our Coarse-to-Fine Quantization.} We take the Tanks\&Temples~\cite{knapitsch2017tanks} dataset as an example to evaluate the effectiveness of different quantization stages in our C2FQ strategy, as shown in Table~\ref{tab:c2f}.
To assess the impact of the quantization matrix, we remove it from the full framework, resulting in the variant “Ours w/o QM.” Compared to the full model, “Ours w/o QM” leads to a 0.01 dB drop in PSNR and a 0.15 MB increase in storage cost.
Furthermore, we remove the quantization vector from “Ours w/o QM,” yielding the variant “Ours w/o QM \& QV,” which causes an additional 0.13 dB PSNR drop and a further 0.23 MB increase in storage cost.
These results validate the effectiveness of both the quantization matrix and vector in enhancing compression efficiency within the proposed C2FQ strategy.


\textbf{Analysis of the Storage Cost.} To further demonstrate the effectiveness of our method, we present a detailed breakdown of the storage cost for each component, as shown in Table~\ref{tab:storage_cost}. The experimental results show that by incorporating both the MoP network and the Coarse-to-Fine Quantization module in the anchor attribute compression process, our method reduces the storage cost of anchor attributes by over 25\% and the total 3DGS compression size by approximately 24\% compared to HAC~\cite{chen2025hac}, while achieving comparable SSIM and even better PSNR and LPIPS scores.
These results highlight the effectiveness of our method in compressing 3DGS data, particularly in optimizing attribute storage.

\begin{figure}[t!]
    \centering
    \includegraphics[width=\linewidth]{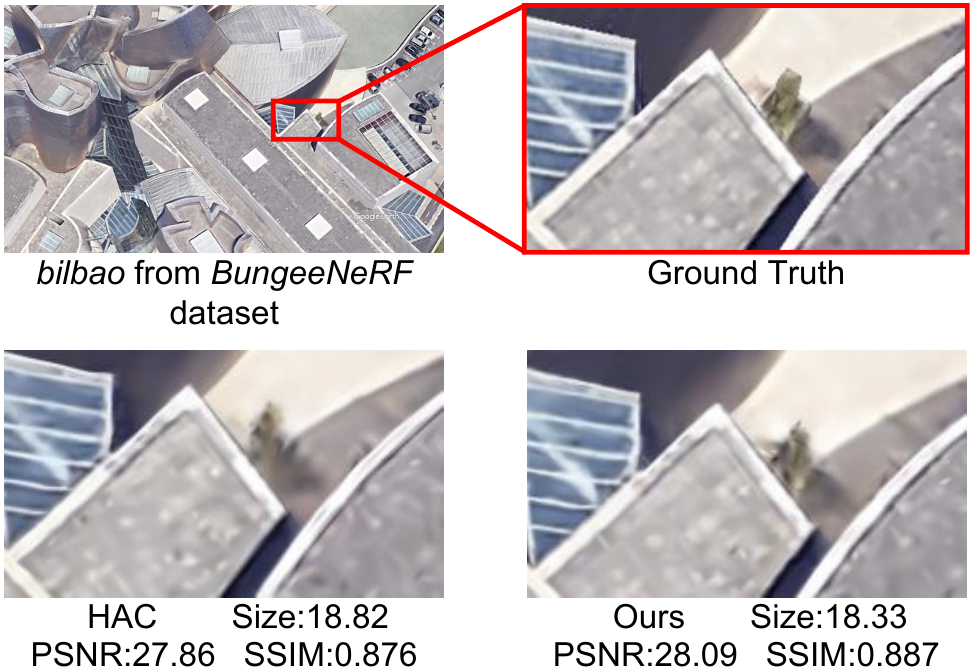}
    \caption{Visualization comparison between HAC and our method on the ``bilbao'' scene from the BungeeNeRF dataset. PSNR, SSIM, and scene size (in MB) of the rendered images are reported.}
    \label{fig:visulization}
\end{figure}

\textbf{Visualization.} We visualize the ``bilbao'' scene from the BungeeNeRF~\cite{xiangli2022bungeenerf} dataset, as shown in Figure~\ref{fig:visulization}. Compared to HAC~\cite{chen2025hac}, the image rendered by our method exhibits more distinct structural details and clearer textures. The quantitative metrics also demonstrate that our method achieves higher fidelity while consuming less storage. These visualization results further validate the effectiveness of our approach. 

\section{Conclusion}


To enhance 3DGS data compression performance, we propose a MoP strategy, which leverages diverse priors generated by multiple lightweight MLPs and combines them using a learnable gating mechanism to produce a unified MoP feature. The resulting feature improves both lossy and lossless compression by serving as enriched hyperprior information for conditional entropy coding and by guiding the Coarse-to-Fine Quantization (C2FQ) procedure to enable element-wise quantization.
Comprehensive experiments demonstrate the effectiveness of our MoP module, achieving state-of-the-art performance in 3DGS compression.
Beyond establishing a strong baseline for efficient 3DGS data compression, our work also inspires future research toward unified modules that jointly address both lossy and lossless compression for 3DGS data.


\newpage
\bibliographystyle{ACM-Reference-Format}
\bibliography{sample-base}




\end{document}